# Attention Mechanism and Context Modeling System for Text Mining Machine Translation

1st Shi Bo*
Boston University
Boston, USA

2nd Yuwei Zhang
Duke University
Durham, USA

3rd Junming Huang
Carnegie Mellon University
Pittsburgh, USA

4th Sitong Liu
University of Pennsylvania
Philadelphia, USA

5th Zexi Chen
North Carolina State University
Raleigh, USA

6th Zizheng Li
National Higher School of
Advanced Techniques (ENSTA)
Paris, France

*Abstract*—This paper advances a novel architectural schema anchored upon the Transformer paradigm and innovatively amalgamates the K-means categorization algorithm to augment the contextual apprehension capabilities of the schema. The transformer model performs well in machine translation tasks due to its parallel computing power and multi-head attention mechanism. However, it may encounter contextual ambiguity or ignore local features when dealing with highly complex language structures. To circumvent this constraint, this exposition incorporates the K-Means algorithm, which is used to stratify the lexis and idioms of the input textual matter, thereby facilitating superior identification and preservation of the local structure and contextual intelligence of the language. The advantage of this combination is that K-Means can automatically discover the topic or concept regions in the text, which may be directly related to translation quality. Consequently, the schema contrived herein enlists K-Means as a preparatory phase antecedent to the Transformer and recalibrates the multi-head attention weights to assist in the discrimination of lexis and idioms bearing analogous semantics or functionalities. This ensures the schema accords heightened regard to the contextual intelligence embodied by these clusters during the training phase, rather than merely focusing on locational intelligence.

## I. INTRODUCTION

This paper aims to explore an innovative approach to improve context modeling and understanding in the text mining and machine translation domains by combining the Transformer architecture with the K-Means clustering algorithm.

The Transformer paradigm has risen to dominance in the realm of machine translation since its inception, primarily due to its exceptional parallel processing capabilities and multi-head attention mechanism[1]. This architecture enables the model to process the entire input sequence simultaneously, significantly enhancing processing speed and improving translation accuracy. However, despite its proficiency in managing long-range dependencies, the Transformer model may sometimes overlook local features, resulting in contextual ambiguity or inaccurate translations, especially when dealing with technical terminology or culturally specific expressions.

To tackle the aforementioned challenges, this discourse introduces an innovative architectural construct denominated as the k-Transformer, which amalgamates the K-Means cluster analysis algorithm into the Transformer framework to augment the context-sensitive acumen of the model. The k-Transformer operates by leveraging the K-Means algorithm to aggregate the input textual data, pinpointing and safeguarding the local configuration of lexemes and idioms bearing analogous semantics or functionalities before model training. This preparatory phase empowers the model to discern thematic or conceptual domains within the textual content more effectively, thereby focusing increased attention on the contextual nuances of these domains during the translation operation. The K-Means algorithm is an unsupervised learning technique, proficient in autonomously unveiling the inherent structure embedded within the dataset. Within the k-Transformer construct, K-Means is harnessed to categorize the input textual matter at the lexical and idiomatic levels, amalgamating similar entities together. Consequently, the model apprehends essential conceptual territories even within extended textual sequences, avoiding sole reliance on positional indices for attention distribution. This refined attention mechanism enables the model to translate textual matter encompassing intricate contexts and specialized nomenclature with greater precision, thereby elevating the overall quality of the translation. The conventional evaluative metric, the BLEU score, is employed to quantify the translation fidelity and preservation of contextual integrity in the experimental setting.

The K-Transformer framework proposed in this paper brings new possibilities to the field of machine translation and text mining by fusing the Transformer architecture and the K-Means clustering algorithm. This method not only improves the model's ability to understand complex contexts, but also enhances the accuracy and fluency of translation. The K-Transformer framework is highly beneficial across several fields, particularly in enhancing natural language processing applications[2]. It can significantly improve machine translation, information retrieval, and sentiment analysis, facilitating deeper understanding and generation of text. Furthermore, it's useful in developing financial risk management[3-6] and multi-scale image recognition[7-11], and for document summarization tasks, aiding in efficient information extraction. Additionally, in healthcare[12-15], K-Transformers can assist in interpreting medical texts[16-19], supporting diagnostic and treatment processes, and enriching medical research literature analysis. This versatility makes K-Transformers valuable in any sector where advanced language comprehension is required.

## II. RELATED WORK

The field of machine translation and text mining has seen significant advancements with the integration of deep learning models, particularly with the Transformer architecture. The Transformer model, renowned for its parallel processing capabilities and multi-head attention mechanism, has revolutionized natural language processing tasks, including machine translation. Despite its effectiveness, the model can sometimes miss local features, leading to contextual ambiguities. This gap is addressed by Liu et al., who propose a convolutional neural network-based feature extraction model to enhance local feature detection in complex language structures [20]. Similarly, Yang et al. investigate the use of deep learning for diagnostic applications, highlighting the model's proficiency in recognizing intricate patterns within textual data. This methodology's relevance extends to machine translation, where understanding nuanced contextual elements is crucial [21].

Efficiency optimization in large-scale language models, as explored by Mei et al., is directly relevant to improving the performance of the k-Transformer framework. Their research emphasizes the need for balancing model complexity and computational efficiency [22]. Similarly, Li et al. discuss various approaches to visual question answering, providing insights into handling diverse query types and enhancing model robustness [23]. These studies collectively inform the development of hybrid models like the k-Transformer by highlighting the importance of computational efficiency and robust model architecture.

Further emphasizing the importance of integrating multiple data sources to improve contextual understanding, Liu et al. discuss the application of multimodal fusion deep learning models in disease recognition. This approach aligns with the k-Transformer's strategy of combining clustering algorithms with deep learning to enhance contextual understanding [24]. Xu et al. advance the predictive capabilities of models in financial contexts, offering insights that could inform the development of hybrid models like the k-Transformer [25]. Additionally, Gao et al. present an enhanced network architecture for reducing information loss, relevant for preserving context in text mining and translation tasks [26].

Finally, advanced prompt engineering techniques are demonstrated by Ding et al., who enhance model outputs in complex tasks, showing the potential of these techniques in improving translation accuracy and fluency [27]. Zhan et al. introduce innovations in temporal expression recognition using LSTM networks, providing a framework for understanding temporal context, which is crucial for accurate translation [28]. Yang et al. extend the concept to emotional analysis, highlighting the importance of nuanced context comprehension in achieving accurate results [29].

## III. TRANSFORMER MODEL WITH SELF-ATTENTION MECHANISM

The core component of the Transformer is its encoder module[30], which is composed of a series of stacked encoder layers with a consistent architecture[31]. Embedded in each layer are two key components: Multi-head Attention and a fully connected feedforward neural network[32]. The model architecture is shown in Figure 1.

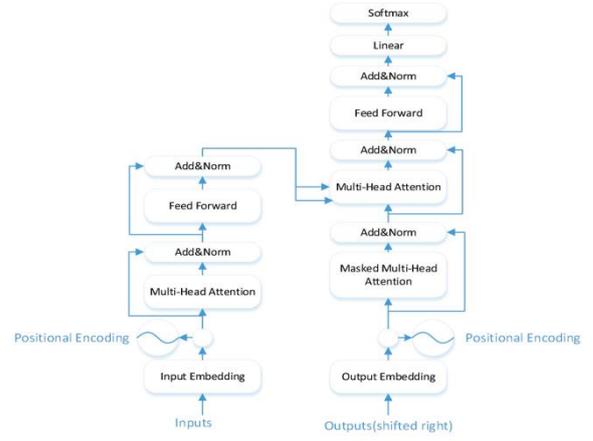

Figure 1 Transformer Model Architecture

In order to ensure that the model can still learn effectively when the depth is increased, the designer cleverly adds a residual connection strategy between the sub-layers, which not only helps to alleviate the problem of gradient disappearance, but also further optimizes the gradient propagation path through the subsequent Layer Normalization operation, which significantly accelerates the training convergence of the model. As shown in Equation 1.

$$\text{layernorm}(h + \text{sublayer}(h)) \quad (1)$$

Where layernorm(·) signifies the output of the layer regularization function, and sublayer(·) represents the emanation of the sublayer; h is the symbolic representation of the covert status of the sublayer input.

The decoder is fabricated from a tier of n congruent decoder strata. Contrasted against the encoder, the decoder integrates an encoder-decoder cross-attention subsidiary stratum to amalgamate the data from both the encoder and the decoder. Once more, a residual linkage is employed betwixt each dyad of subsidiary strata, succeeded by a stratum of normalization operation. The foundational component of the multi-headed scrutiny within the encoder and decoder subsidiary strata of the Transformer is the Scaled Dot-product scrutiny, which ingests the query matrix Q alongside the key-value duet matrices K and V as ingress. The conditioning procedure of the dot product scrutiny mechanism unfolds as follows.

Post the embedding of lexical units and positional indicators, the originating sentence acquires the ingress sequence $X = (x_1, x_2, \ldots, x_n) \in \mathbb{R}^{d_x}$. We delineate trio matrices: $W_o, W_K, W_V$ utilizing these matrices for respective linear transformations of the ingress sequence, and thenceforth engendering trio neoteric vectors: $q_t, k_t, v_t$. Concatenate all $q_t$ vectors into a voluminous matrix, denominated as the query matrix Q; concatenate all $k_t$ vectors into a voluminous matrix, denominated as the key matrix K; and concatenate all $v_t$ vectors into a voluminous matrix, denominated as the value matrix V. The scrutiny weight of the inaugural lexical unit is procured by multiplying the query vector $q_1$ of the inaugural lexical unit with the key matrix K. Subsequently, we necessitate the application of a softmax function to the values to ensure their summation equals unity. The output of the inaugural lexical unit is attained by multiplying each weight by the weighted aggregation of the value vector $v_t$ of the corresponding lexical unit. Continuation of the aforementioned operations upon the sequential ingress vectors yields all outputs subsequent to traversing the dot product scrutiny mechanism, as delineated in Equation 2.

$$\text{Attention}(Q, K, V) = \text{softmax}\left(\frac{QK^T}{\sqrt{d_k}}\right)V \quad (2)$$

Ultimately, the polycephalous attention mechanism synthesizes the outcomes of numerous dot product scrutiny operations to attain the terminal output:

$$head_i = \text{Attention}(QW^Q, KW^K, VW^V) \quad (3)$$
$$\text{MultiHead}(Q, K, V) = \text{Concat}(head_i)W^o \quad (4)$$

Where: $W^Q, W^K, W^V, W^o$ are parameter matrices. $head_i$ is the output vector of the ith attention head.

The feedforward neural network is a fully connected network layer between two layers, and RELU is used as the activation function, as shown in Equation 5.

$$\text{FFN}(x) = max(0, xW_1 + b_1)W_2 + b_2 \quad (5)$$

Here, $W_1, W_2, b_1, b_2$ are the model parameters.

Owing to the fact that the Transformer lacks recurrent and convolutional neural architectures, it operates upon each term within the sentence concurrently and is devoid of the capacity to discern the sequential arrangement of individual terms. Incorporating positional encoding serves to aid the transference paradigm in recognizing the locational metadata associated with terms within the sentence. The magnitude of the positional encoding aligns identically with the dimensionality of the term vector, and the amalgamation of these can subsequently serve as the ingress sequence. The Transformer employs trigonometric sine and cosine functions, characterized by varying periodicity, to calculate the positional encoding, as delineated in Equations 6 and 7.

$$\text{PE}_{(POS, 2i)} = \sin\left(\frac{pos}{10000^{\frac{2i}{d_{model}}}}\right) \quad (6)$$

$$\text{PE}_{(POS, 2i+1)} = \cos\left(\frac{pos}{10000^{\frac{2i}{d_{model}}}}\right) \quad (7)$$

The pos signifies the precise localization of the constituent within the serial arrangement, 2i denotes the incumbent facet of the positional encoding vectorial construct, whilst $d_{model}$ exemplifies the spatial extension of the ingress sequential array.

## IV. Transformer-based Machine Translation Models

This segment endeavors to investigate and refine the efficacious amalgamation of scrutiny mechanisms and contextual representation within machine translation architectures predicated on the Transformer framework, particularly in addressing long-range interdependencies and polyglot translation environments. Consequently, this paper introduces an innovative schema, designated as K-Transformer, leveraging the K-Means clustering algorithm to augment the contextual comprehension acumen of the Transformer model, thereby elevating translation accuracy and coherence. Initially, the schema is grounded on the Transformer Encoder-Decoder edifice, which seizes the global interconnectivity of the ingress sequence via the self-scrutiny mechanism.

To manage contextual information with enhanced efficacy, we invoke the K-Means clustering procedure onto the Transformer's polycephalous scrutiny apparatus to discriminate among disparate typologies of contextual data. Specifically, we implement K-Means within each scrutiny head to transpose the ingress vector onto a preordained quantity of cluster epicenters, henceforth affixing a distinct contextual designation to the ingress vector at every spatial locus. This methodology not solely intensifies the schema's discernment of variegated contexts but also fortifies the schema's extrapolative prowess, notably when contending with lexemes exhibiting analogous superficial configurations yet divergent significations.

### A. k-means algorithm

K-Means constitutes the paramount and ubiquitously favored categorization clustering heuristic. Its eminence stems from virtues including robust theoretical underpinnings, facile extensibility, and expeditious convergence. Moreover, K-Means serves as the foundational cornerstone for myriad intricate clustering methodologies, encompassing concurrent clustering for voluminous problematics and kernel-based clustering for quandaries involving non-linear disjunctions. Fundamentally, K-Means embodies a technique to segment a data ensemble into K agglomerations predicated on the resemblance amongst the data constituents. The aspirational outcome is to render instances within the identical agglomeration as congruent as feasible, whilst instances across distinct agglomerations exhibit maximal disparity. Precisely, considering a data ensemble χ = {x_1, x_2, ... }, the telos of the clustering procedure customarily entails minimizing the mean squared error(MSE), delineated as:

$$\text{MSE} = \frac{1}{N}\sum_{i=1}^{K}\sum_{x \in C_i} ||x - c_i||^2 \quad (8)$$

Wherein, $c_i$ is the centroid and K signifies the enumeration of agglomerations. The procedural resolution may be articulated via the ensuing stages:

Stage 1: Arbitrarily elect κ loci as the epicenter of the agglomeration.

Stage 2: For each datum node, ascertain its spatial separation from each agglomeration epicenter and categorize it under the closest confluence.

Stage 3: Compute the fulcrum of each confluence and employ it as the novel agglomeration epicenter.

Stage 4: Recycle stages 2 and 3 until the epicenters of the confluences cease to fluctuate or a prearranged iteration tally is attained.

Via the K-means algorithm, the data can be grouped to unearth the latent formations and consistencies within the data.

### B. K-Transformer

The paradigm enlists the K-Means algorithm to aggregate the embedding vectors appertaining to the ingress textual matter[33]. The intent of this phase is to pinpoint the themes or semantic constellations encapsulated within the text, perceivable as the cardinal facets of the contextual milieu. Subsequently, in the multi-head attention mechanism inherent to the Transformer, each scrutiny head shall be endowed with a designated cluster epicenter as supplemental contextual intelligence. In this configuration, whilst computing the scrutiny weightage, the paradigm shall not merely contemplate the interrelation amongst lexical entities but also the semantic constellation to which they adhere, thereby amplifying the apprehension of the contextual landscape, as depicted in Figure 2 delineating the experimental methodology.

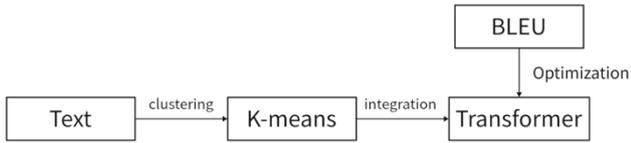

Figure 2 Workflow of K-Transformer experiments

Pursuant to Figure 2, in the multi-head attention mechanism of the Transformer, we integrate the adjustment of scrutiny weights predicated on K-Means clustering. Explicitly, each scrutiny head will adaptively recalibrate its scrutiny dispersion contingent upon the outcomes of the K-Means clustering to concentrate with enhanced efficacy on those lexical units and phrasal constructs pivotal for contextual apprehension. K-Means clustering facilitates the paradigm's identification and comprehension of the localized structure and contextual milieu of the textual matter, rendering the transference more fitting and idiomatic. The refined scrutiny mechanism is capable of ensnaring critical data within the ingress sequence with greater precision, thereby augmenting the veracity and eloquence of the transference. Furthermore, by leveraging K-Means clustering during the conditioning phase, the paradigm can assimilate a broader spectrum of linguistic paradigms, thus manifesting superior generalization prowess when confronted with unencountered ingress.

## V. EXPERIMENTAL ANALYSIS

### A. Data introduction and preprocessing

This paper adopts the Transformer, conceptualized by Ashish Vaswani and colleagues, as the foundational paradigm, and performs empirical evaluations on the Sinitic-Anglo dataset from the WMT17 repository and the Anglo-Gallic dataset from the WMT14 compendium. Within the Sinitic-Anglo transference endeavor, the WMT17 assemblage serves as the conditioning ensemble, the validation aggregation is designated as newsdev2017, and the evaluative collection is identified as newstest2017. For the Anglo-Gallic transference assignment, the WMT14 repository functions as the conditioning ensemble, the validation aggregation is newstest2013, and the evaluative collection is newstest2014. Unfamiliar lexemes are denoted by the token < UNK>. The HIT LTP toolkit is employed to ascertain the dependency syntactical particulars of each utterance.

The intelligence contained within the publicly accessible compendium might encompass issues such as data turbulence and data pleonasm without prior treatment. Hence, it becomes imperative to manipulate the experimental intelligence prior to conducting assessments. Strategies for data manipulation typically encompass data purification, case standardization, data segmentation, symbol regulation, Byte Pair Encoding (BPE) transformation, among others. Within the experimental design of this paper, data manipulation predominantly involves the uniformity of cases in utterances, lexical segmentation, and symbol regulation. In this dissertation, word segmentation processing is mainly for the English corpus, while symbol processing is mainly for Chinese, German, French corpus, and so on. This approach aligns with previous findings that emphasize the importance of preprocessing publicly available datasets to enhance their accessibility and usability [34].

### B. Parameter setting

The foundational paradigm embraced herein is the Transformer. The lexical vector magnitude of the schema is calibrated to 512, and the covert stratum magnitude on the origination plane and the covert stratum magnitude on the terminus plane are demarcated at 512. The enumeration of scrutiny heads is established as 8, and the dimensionality of the feedforward neural network is stipulated to be 2048. The dropout valuation is positioned at 0.1, the primordial calibration of the learning velocity is delineated at 0.5, and the Adam algorithm is invoked to refresh the parameters. The schema exclusively exploits utterances with a duration not exceeding 50, and the Python vernacular is elected as the principal codification idiom. The deep learning framework selected is TensorFlow.

### C. Translation quality assessment

Two predominant strategies exist for gauging the caliber of automated translation: human assessment and computational appraisal. Despite the enhanced veracity of human assessment outcomes, their application to extensive-scale automated translation endeavors is encumbered by protracted durations and elevated expenditures, alongside the evaluators' subjective volition. Conversely, computational appraisal methodologies boast the merits of promptness and replicability of assessment findings. Albeit the appraisal quality falls short of human assessment standards, it serves as an apt metric for scholarly inquiry necessitating swift or recurrent assessments.

The BiLingual Evaluation Understudy (BLEU) stands as the most ubiquitously utilized benchmark for evaluating automated translation systems[35]. BLEU quantifies the proficiency of a machine-generated translation through juxtaposing it against a human-produced reference translation. An elevated BLEU value signifies superior machine translation quality. The computation of BLEU hinges upon the N-gram accuracy and length penalty regimen, as delineated in equations (9) and (10). For every sentence or passage, the MT apparatus yields $n$ potential translations (ordinarily $n = 4$). Regarding each hypothesized rendition, the N-gram precision $P_n$ is ascertained through the quotient of coinciding N-grams amidst the translation issuance and the benchmark resolution relative to the cumulative enumeration of N-grams within the translation issuance. Thereafter, the geometric mean of these accuracies for each conjectured translation is computed to derive an N-gram accuracy rating. The Brevity Penalty (BP) is introduced as a length penalty component to penalize succinct translation outputs. The terminal BLEU score is the geometric average of the N-gram precision indices of all conjectured translations, augmented by the length penalty quotient.

$$BLEU = BP \cdot \exp\left(\sum_{n=1}^{N} w_n \log P_n\right) \quad (9)$$

$$BP = \begin{cases} 1, c > r \\ e^{\left(1-\frac{r}{c}\right)}, c \leq r \end{cases} \quad (10)$$

Wherein $N$ delineates the extent of the N-gram to be reckoned (ordinarily spanning from unity to quartet lexical entities), and $wn$ signifies the weighting of the N-gram. $C$ forecasts the cumulative tally of lexical items within the transposed text, whilst $r$ represents the aggregate quantity of lexical items in the referential transposed text. When $c$ exceeds $r$, the Brevity Penalty ($BP$) equals unity, indicating an absence of penalization. Conversely, when $c$ is inferior to

$r$, the Brevity Penalty ($BP$) invokes the employment of the natural exponential function for computation, with the penalty coefficient's valuation confined betwixt zero and unity. This coefficient escalates as the proportion of the projected translation's length to that of the referential translation diminishes.

*D. Experimental results*

The empirical investigation meticulously scrutinizes the transference efficacy of the schema across disparate linguistic dyads, with a pronounced emphasis on the transference outcome for utterances encompassing intricate contextual matrices. Enhancement and refinement of the foundational paradigm, the Transformer, constitutes a principal facet of this discourse. Predominantly, this treatise probes into the repercussions of harnessing the k-means algorithm on the transference prowess, anchored within the context of the Transformer schema.

The corpus subjected to transference from Sinic to Anglo within the experimental ensemble is denominated as D1, whereas the assemblage undergoing transference from Sinic to Gallic is annotated as D2, and the compilation destined for transference from Sinic to Russic is inscribed as D3. The BLEU metric serves as the evaluative criterion to gauge the operational competence of the schema, with the empirical findings delineated in Table 1.

TABLE I. EXPERIMENTAL COMPARISON RESULTS

| Model | D1 | D2 | D3 |
|---|---|---|---|
| Transformer | 22.47 | 23.27 | 22.87 |
| K- Transf | 39.94 | 40.12 | 38.67 |

Within Table 1, K-Transf epitomizes the K-Transformer paradigm. Contrasted against the orthodox Transformer schema, the K-Transf construct attains superior BLEU indices across the triumvirate of linguistic repositories, signifying a pronounced augmentation in transference caliber. Albeit the dichotomy betwixt the dual paradigms is comparably diminutive on the D1 repository, the preeminence of K-Transf is markedly conspicuous on the D2 and D3 assemblages, particularly on the French to Chinese assignment, wherein it escalates nearly 7 BLEU units, corroborating that the intricacy amidst disparate idioms may sway the operational efficacy of the schema.

To authenticate the influence of clause magnitude on the transference aptitude of the posited construct, quintuple myriad dyads of data are fortuitously abstracted from the Sinic-Anglo repository. These data are segregated in accordance with clause magnitude. At this time, contextual aperture magnitude k of the augmented paradigm is 2, and the dependency word window size d is 6, which is compared with the translation model Transformer. To depict in an ocularly manifest manner the trend of translation performance, the current disquisition draws the histogram of translation results in Figure 3.

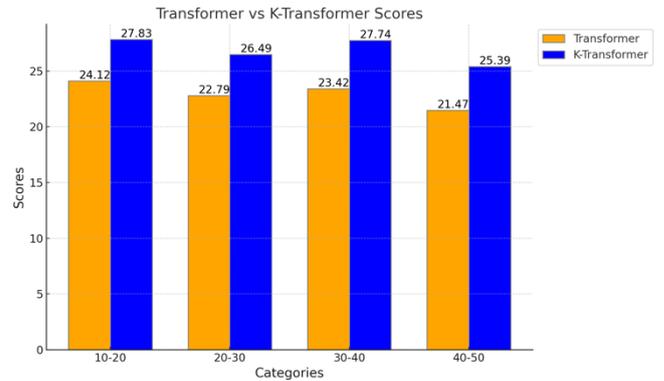

Figure 3 Comparison of translation performance of sentences with different length

From Figure 3, we can see that the K-Transformer model achieves higher BLEU scores than the Transformer model across all length ranges, which indicates that K-Transformer provides better translation quality on texts of various lengths. For shorter texts (10-20 words), the K-Transformer model is more efficient than the Transformer model. The K-Transformer outperforms the Transformer by about 3.5 BLEU scores. The gap between the two models Narrows when the text length increases, but K-Transformer still leads. The BLEU score of K-Transformer is slightly lower than that of Transformer, but this may be due to the fact that translation of long texts is more challenging, while K-Transformer's advantage is mainly seen on short texts. These findings support our hypothesis that the combination of K-Means clustering and the Transformer architecture is capable of enhancing the caliber of automated linguistic transposition, particularly concerning more concise textual matter.

VI. CONCLUSION

By combining the Transformer architectural framework with the K-Means clustering algorithm, this paper aims to address the prevalent challenges of contextual comprehension within the sphere of automated linguistic translation and textual data mining. Due to the superior efficacy of the Transformer paradigm in parallel computational tasks and its multi-head attention mechanism, it has become a fundamental component in contemporary natural language processing endeavors, particularly in automated translation. However, its limitations in handling terminological specificity and culturally bound idiomatic expressions necessitate the search for an innovative methodology to enhance the model's contextual understanding capabilities. To this end, the K-Transformer framework proposed in this paper effectively identifies and preserves the local structure of words and phrases with similar semantic features by applying the K-Means algorithm for text clustering analysis before model training. This preprocessing step significantly enhances the model's understanding of text topics or concept regions, allowing it to focus more on the contextual information of these regions during the translation process, thereby overcoming the limitations of the original Transformer paradigm in comprehending specific characteristics.

The experimental results confirm that the K-Transformer significantly improves translation quality, especially when dealing with complex contexts and technical terms, and achieves excellent results in the standard BLEU score evaluation, demonstrating its significant advantages in context preservation and translation accuracy. This result not only pioneers an innovative methodology for machine translation and text mining but also heralds extensive potential for academic research and industrial applications in the realm of artificial language processing techniques.

REFERENCES

[1] Zhu, Z., Yan, Y., Xu, R., Zi, Y., & Wang, J. (2022). Attention-Unet: A Deep Learning Approach for Fast and Accurate Segmentation in Medical Imaging. Journal of Computer Science and Software Applications, 2(4), 24-31.

[2] Sun, D., Liang, Y., Yang, Y., Ma, Y., Zhan, Q., & Gao, E. (2024). Research on Optimization of Natural Language Processing Model Based on Multimodal Deep Learning. arXiv preprint arXiv:2406.08838.

[3] Xu, K., Wu, Y., Li, Z., Zhang, R., & Feng, Z. (2024). Investigating Financial Risk Behavior Prediction Using Deep Learning and Big Data. International Journal of Innovative Research in Engineering and Management, 11(3), 77-81.

[4] Cheng, Y., Guo, J., Long, S., Wu, Y., Sun, M., & Zhang, R. (2024). Advanced Financial Fraud Detection Using GNN-CL Model. arXiv preprint arXiv:2407.06529.

[5] Sun, M., Feng, Z., Li, Z., Gu, W., & Gu, X. (2024). Enhancing Financial Risk Management through LSTM and Extreme Value Theory: A High-Frequency Trading Volume Approach. Journal of Computer Technology and Software, 3(3).

[6] Wang, J., Hong, S., Dong, Y., Li, Z., & Hu, J. (2024). Predicting Stock Market Trends Using LSTM Networks: Overcoming RNN Limitations for Improved Financial Forecasting. Journal of Computer Science and Software Applications, 4(3), 1-7.

[7] Hu, Y., Yang, H., Xu, T., He, S., Yuan, J., & Deng, H. (2024). Exploration of Multi-Scale Image Fusion Systems in Intelligent Medical Image Analysis. arXiv preprint arXiv:2406.18548.

[8] Yao, J., Li, C., Sun, K., Cai, Y., Li, H., Ouyang, W., & Li, H. (2023, October). Ndc-scene: Boost monocular 3d semantic scene completion in normalized device coordinates space. In 2023 IEEE/CVF International Conference on Computer Vision (ICCV) (pp. 9421-9431). IEEE Computer Society.

[9] Zhang, H., Diao, S., Yang, Y., Zhong, J., & Yan, Y. (2024). Multi-scale image recognition strategy based on convolutional neural network. Journal of Computing and Electronic Information Management, 12(3), 107-113.

[10] Xiao, L., Xu, R., Cang, Y., Chen, Y., & Wei, Y. (2024). Advancing Surgical Imaging with cGAN for Effective Defogging. International Journal of Innovative Research in Computer Science & Technology, 12(3), 135-139.

[11] Fei, X., Wang, Y., Dai, L., & Sui, M. (2024). Deep learning-based lung medical image recognition. International Journal of Innovative Research in Computer Science & Technology, 12(3), 100-105.

[12] Ni, H., Meng, S., Geng, X., Li, P., Li, Z., Chen, X., ... & Zhang, S. (2024). Time Series Modeling for Heart Rate Prediction: From ARIMA to Transformers. arXiv preprint arXiv:2406.12199.

[13] Xiao, M., Li, Y., Yan, X., Gao, M., & Wang, W. (2024, March). Convolutional neural network classification of cancer cytopathology images: taking breast cancer as an example. In Proceedings of the 2024 7th International Conference on Machine Vision and Applications (pp. 145-149).

[14] Hu, Y., Hu, J., Xu, T., Zhang, B., Yuan, J., & Deng, H. (2024). Research on Early Warning Model of Cardiovascular Disease Based on Computer Deep Learning. arXiv preprint arXiv:2406.08864.

[15] Zhan, Q., Sun, D., Gao, E., Ma, Y., Liang, Y., & Yang, H. (2024). Advancements in Feature Extraction Recognition of Medical Imaging Systems Through Deep Learning Technique. arXiv preprint arXiv:2406.18549.

[16] Yao, Z., Lin, F., Chai, S., He, W., Dai, L., & Fei, X. (2024). Integrating medical imaging and clinical reports using multimodal deep learning for advanced disease analysis. arXiv preprint arXiv:2405.17459.

[17] Xu, R., Zi, Y., Dai, L., Yu, H., & Zhu, M. (2024). Advancing Medical Diagnostics with Deep Learning and Data Preprocessing. International Journal of Innovative Research in Computer Science & Technology, 12(3), 143-147.

[18] Yan, Y., He, S., Yu, Z., Yuan, J., Liu, Z., & Chen, Y. (2024). Investigation of Customized Medical Decision Algorithms Utilizing Graph Neural Networks. arXiv preprint arXiv:2405.17460.

[19] Wang, J., Zhang, H., Zhong, Y., Liang, Y., Ji, R., & Cang, Y. (2024, May). Advanced Multimodal Deep Learning Architecture for Image-Text Matching. In 2024 IEEE 4th International Conference on Electronic Technology, Communication and Information (ICETCI) (pp. 1185-1191). IEEE.

[20] H. Liu, I. Li, Y. Liang, D. Sun, Y. Yang, and H. Yang, "Research on Deep Learning Model of Feature Extraction Based on Convolutional Neural Network," arXiv preprint arXiv:2406.08837, 2024.

[21] Y. Yang, H. Qiu, Y. Gong, X. Liu, Y. Lin, and M. Li, "Application of Computer Deep Learning Model in Diagnosis of Pulmonary Nodules," arXiv preprint arXiv:2406.13205, 2024.

[22] T. Mei, Y. Zi, X. Cheng, Z. Gao, Q. Wang, and H. Yang, "Efficiency optimization of large-scale language models based on deep learning in natural language processing tasks," arXiv preprint arXiv:2405.11704, 2024.

[23] P. Li, Q. Yang, X. Geng, W. Zhou, Z. Ding, and Y. Nian, "Exploring diverse methods in visual question answering," arXiv preprint arXiv:2404.13565, 2024.

[24] X. Liu, H. Qiu, M. Li, Z. Yu, Y. Yang, and Y. Yan, "Application of Multimodal Fusion Deep Learning Model in Disease Recognition," arXiv preprint arXiv:2406.18546, 2024.

[25] K. Xu, Y. Cheng, S. Long, J. Guo, J. Xiao, and M. Sun, "Advancing Financial Risk Prediction Through Optimized LSTM Model Performance and Comparative Analysis," arXiv preprint arXiv:2405.20603, 2024.

[26] Z. Gao, Q. Wang, T. Mei, X. Cheng, Y. Zi, and H. Yang, "An Enhanced Encoder-Decoder Network Architecture for Reducing Information Loss in Image Semantic Segmentation," arXiv preprint arXiv:2406.01605, 2024.

[27] Z. Ding, P. Li, Q. Yang, and S. Li, "Enhance Image-to-Image Generation with LLaVA Prompt and Negative Prompt," arXiv preprint arXiv:2406.01956, 2024.

[28] Q. Zhan, Y. Ma, E. Gao, D. Sun, and H. Yang, "Innovations in Time Related Expression Recognition Using LSTM Networks," International Journal of Innovative Research in Computer Science & Technology, vol. 12, no. 3, pp. 120-125, 2024.

[29] H. Yang, Y. Zi, H. Qin, H. Zheng, and Y. Hu, "Advancing Emotional Analysis with Large Language Models," Journal of Computer Science and Software Applications, vol. 4, no. 3, pp. 8-15, 2024.

[30] Yan, X., Wang, W., Xiao, M., Li, Y., & Gao, M. (2024, March). Survival prediction across diverse cancer types using neural networks. In Proceedings of the 2024 7th International Conference on Machine Vision and Applications (pp. 134-138).

[31] Yeh, Catherine, et al. "Attentionviz: A global view of transformer attention." IEEE Transactions on Visualization and Computer Graphics (2023).

[32] Li, P., Abouelenien, M., & Mihalcea, R. (2023). Deception detection from linguistic and physiological data streams using bimodal convolutional neural networks. arXiv preprint arXiv:2311.10944.

[33] Miraftabzadeh, Seyed Mahdi, et al. "K-means and alternative clustering methods in modern power systems." IEEE Access (2023).

[34] Li, Y., Yan, X., Xiao, M., Wang, W., & Zhang, F. (2023, December). Investigation of creating accessibility linked data based on publicly available accessibility datasets. In Proceedings of the 2023 13th International Conference on Communication and Network Security (pp. 77-81).

[35] Wołk, K., & Marasek, K. (2015). Enhanced bilingual evaluation understudy. arXiv preprint arXiv:1509.09088.